\documentclass[11pt,a4paper]{article}
\usepackage[hyperref]{acl2020}
\usepackage{times}
\usepackage{latexsym}

\usepackage{graphicx}
\usepackage{comment}
\usepackage{times}
\usepackage{latexsym}
\usepackage{multirow}
\usepackage{soul}
\usepackage{color}
\usepackage{nccmath}

\usepackage{graphicx}
\usepackage{caption}
\usepackage{subcaption}

\soulregister\ref{7}
\soulregister\citep{7}
\soulregister\citep{7}
\soulregister\citet{7}
\soulregister\textbf{7}
\soulregister\textit{7}

\usepackage{makecell}
\usepackage{enumitem}
\usepackage{lipsum}
\usepackage{blindtext}

\usepackage{url}

\usepackage{microtype}

\aclfinalcopy 

\title{Extensive Error Analysis and a Learning-Based Evaluation of Medical Entity Recognition Systems to Approximate User Experience}

\author{Isar Nejadgholi, Kathleen C. Fraser and Berry De Bruijn\\
National Research Council Canada \\
\texttt \footnotesize \{isar.nejadgholi, kathleen.fraser,berry.debruijn\}@nrc-cnrc.gc.ca}

\date{}

\begin{document}
\maketitle
\begin{abstract}
When comparing entities extracted by a medical entity recognition system with gold standard annotations over a test set, two types of mismatches might occur, label mismatch or span mismatch. Here we focus on span mismatch and show that its severity can vary from a serious error to a fully acceptable entity extraction due to the subjectivity of span annotations. For a domain-specific BERT-based NER system, we showed that 25\% of the errors have the same labels and overlapping span with gold standard entities. We collected expert judgement which shows more than 90\% of these mismatches are accepted or partially accepted by the user. Using the training set of the NER system, we built a fast and lightweight entity classifier to approximate the user experience of such mismatches through accepting or rejecting them. The decisions made by this classifier are used to calculate a learning-based F-score which is shown to be a better approximation of a forgiving user's experience than the relaxed F-score. We demonstrated the results of applying the proposed evaluation metric for a variety of deep learning medical entity recognition models trained with two datasets.  
\end{abstract}

\section{Introduction}
\label{sec:intro}


Named entity recognition (NER) in medical texts involves the automated recognition and classification of relevant medical/clinical entities, and has numerous applications including information extraction from clinical narratives \cite{Meystre2008}, identifying potential drug interactions and adverse affects \cite{Harpaz2014,Liu2016}, and de-identification of personal health data \cite{Dernoncourt2017}.

In recent years, medical NER systems have improved over previous baseline performance by incorporating developments such as deep learning models \cite{yadav2018survey}, contextual word embeddings \cite{zhu2018clinical,si2019enhancing}, and domain-specific word embeddings \cite{alsentzer-etal-2019-publicly,10.1093/bioinformatics/btz682,peng2019transfer}. Typically, research groups report their results using common evaluation metrics (most often precision, recall, and F-score) on standardized data sets. While this facilitates exact comparison, it is difficult to know whether modest gains in F-score are associated with significant qualitative differences in the system performance, and how the benefits and drawbacks of different embedding types are reflected in the output of the NER system.


This work aims to investigate the types of errors and their proportion in the output of modern deep learning models for medical NER. We suggest that an evaluation metric should be a close reflection of what users experience when using the model. We investigate different types of errors that are penalized by exact F-score and identify a specific error type where there is high degrees of disagreement between the human user experience and what exact F-score measures: namely, errors where the extracted entity is correctly labeled, but the span only overlaps with the annotated entity rather than matching perfectly. We obtain expert human judgement for 5296 such errors, ranking the severity of the error in terms of end user experience. We then compare the commonly used F-score metrics with human perception, and investigate if there is a way to automatically analyze such errors as part of the system evaluation. The code that calculates the number of different types of errors given the predictions of an NER model and the corresponding annotations is available upon request and will be released at \url{https://github.com/nrc-cnrc/NRC-MedNER-Eval} after publication. We will also release the collected expert judgements so that other researchers can use it as a benchmark for further investigation about this type of errors.

\section{What do NER Evaluation Metrics Measure?}


An output entity from an NER system can be incorrect for two reasons: either the span is wrong, or the label is wrong (or both). Although entity-level exact F-score (also called strict F-score) is established as the most common metric for comparing NER models, exact F-score is the least forgiving metric in that it only credits a prediction when both the span and the label exactly match the annotation.    

Other evaluation metrics have been proposed. The Message Understanding Conference (MUC) used an evaluation which took into account different types of errors made by the system \cite{chinchor93}. Building on that work, the SemEval 2013 Task 9.1 (recognizing and labelling pharmacological substances in biomedical text) employed four different evaluations: \textit{strict match}, in which label and span match the gold standard exactly, \textit{exact boundary match}, in which the span boundaries match exactly regardless of label, \textit{partial boundary match}, in which the span boundaries partially match regardless of label, and \textit{type match}, in which the label is correct and the span overlaps with the gold standard \cite{Segura2013}. The latter metric, also commonly known as \textit{inexact match}, has been used to compute inexact or relaxed F-score in the i2b2 2010 clinical NER challenge \cite{Uzuner2011}. Relaxed F-score and exact F-score are the most frequently used evaluation metrics for measuring the performance of medical NER systems \cite{yadav2018survey}. Other biomedical NER evaluations have accepted a span as a match as long as either the right or left boundary is correct \cite{Tsai2006}. In BioNLP shared task 2013, the accuracy of the boundaries is relaxed or measured based on similarity of entities \cite{bossy2013bionlp}. Another strategy is to annotate all possible spans for an entity and accept any matches as correct \cite{Yeh2005}, although this detailed level of annotation is rare.

Here, we focus on the differences between what F-score measures and the user experience. In the case of a correct label with a span mismatch, it is not always obvious that the user is experiencing an error, due to the subjectivity of span annotations \cite{Tsai2006,kipper2008system}. Existing evaluation metrics treat all such span mismatches equally, either penalizing them all (exact F-score), rewarding them all (relaxed F-score), or based on oversimplified rules that do not generalize across applications and data sets. We use both human judgement and a learning-based approach to evaluate span mismatch errors and the resulting gap between what F-score measures and what a human user experiences. We only consider the information extraction task and not any specific downstream task.


\begin{figure*}[h!]
\centering
\includegraphics[width=1\linewidth]{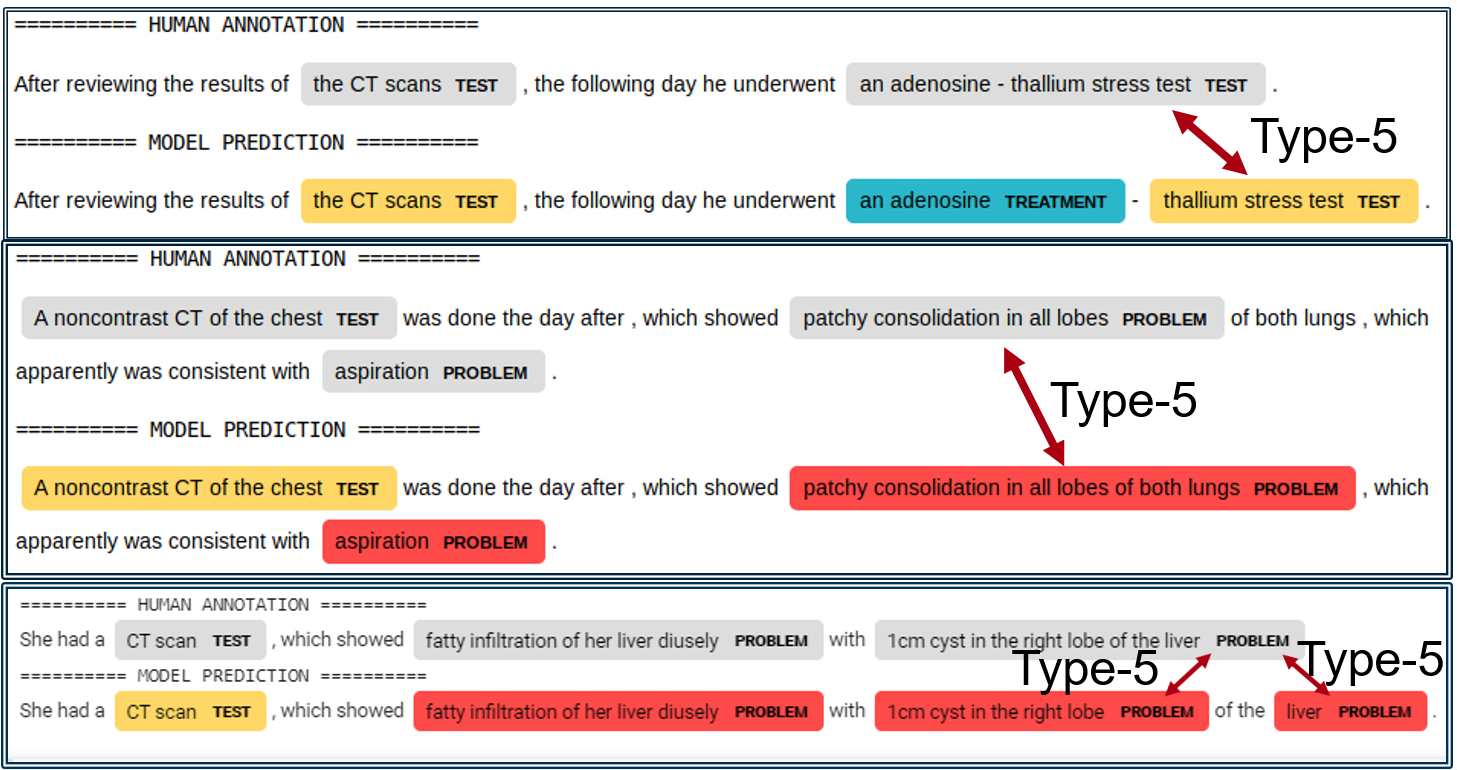}
\caption{Examples of Type-5 error. We used the visualisation tool developed in \cite{zhu2018clinical}}
\label{fig:examples}
\end{figure*}

\section{Types of Errors in NER systems}
\label{sec:types}

While the SemEval 2013 Task 9.1 categorized different types of matches for the purpose of evaluation, we further categorize mismatches for the sake of error analysis. We consider five types of mismatches between annotation and prediction of the NER system. Reporting and comparing the number of these mismatches alongside an averaged score such as F-score can shed light on the differences of NER systems. 

\begin{itemize}
\item \textbf{Mismatch Type-1, Complete False Positive}: An entity is predicted by the NER model, but is not annotated in the hand-labelled text. 
\item \textbf{Mismatch Type-2, Complete False Negative}: A hand labelled entity is not predicted by the model. 
\item \textbf{Mismatch Type-3, Wrong label, Right span}: A hand-labelled entity and a predicted one have the same spans but different tags. 
\item \textbf{Mismatch Type-4, Wrong label, Overlapping span}: A hand-labelled entity and a predicted one have overlapping spans but different tags. 
\item \textbf{Mismatch Type-5, Right label, Overlapping span}: A hand-labelled entity and a predicted one have overlapping spans and the same tags. 
\end{itemize}

We focus on Type-5 errors and show that treating these mismatches is not a trivial task. Previous works have shown that some Type-5 mismatches are completely wrong predictions while others are fully acceptable predictions resulting from the subjectivity and inconsistency of span annotations \cite{Tsai2006}. 


Figure \ref{fig:examples} shows several examples of error Type-5. In the first example, \textit{an adenosine - thallium stress test} is annotated as a \textit{test}, while the NER system extracts \textit{thallium stress test} as a \textit{test}. Here, what NER extracted is partially correct but misses an important part of the entity. Whether the extracted entity is acceptable may depend on the downstream task.
In the next sentence,  \textit{patchy consolidation in all lobes} is annotated as a \textit{problem}, but the NER system extracted \textit{patchy consolidation in all lobes of both longs} as the \textit{problem}. Here, the system's prediction is more complete than the annotated entity, and so it appears to be a fully acceptable prediction. In the last example, according to human annotation, \textit{1cm cyst in the right lobe of the liver} is a \textit{problem}, but the NER system extracts two entities from the same phrase, 1) \textit{1cm cyst in the right lobe} as a \textit{problem} and 2) \textit{liver} as another \textit{problem}. While the first extracted entity is correct and may be acceptable the second one is completely wrong. 


\section{Datasets and Models}
\label{sec:models}

We consider two medical text datasets, one clinical and the other biomedical. We analyse the errors of three models for each dataset to cover a variety of deep learning models.

\paragraph{i2b2 dataset:}

The i2b2 dataset of annotated clinical notes was introduced by \citep{Uzuner2011} in a shared task on entity recognition and relation extraction. The texts, consisting of de-identified discharge summaries, have been annotated for three entity types: problems, tests, and treatments. There are two versions of this dataset, as the version that was released to the wider NLP community contains fewer texts than in the original shared task. We use the second version, which has become an important benchmark in the literature on clinical NER \cite{bhatia2019joint,zhu2018clinical}. There are  170 documents (16520 entities) in the i2b2 train set and 256 documents (31161 entities) in its test set. 

The i2b2 dataset was annotated by community annotators with carefully crafted guidelines. The ground truth generated by the community obtained F-measures above 0.90 against the ground truth of the experts \citep{Uzuner2011}.

\paragraph{MedMentions dataset:}

The MedMentions dataset was released in 2019 and contains 4,392 abstracts from biomedical articles on PubMed \cite{Mohan2019}. The abstracts are annotated for UMLS concepts and semantic types. The fully annotated dataset contains 127 semantic types and these classes are highly-imbalanced. The creators of the dataset also provide a version which has been annotated with only a subset of the most relevant concepts, called `st21pv' (\textit{21 semantic types from preferred vocabularies}); we consider this version in the current work. While fewer papers have been published on MedMentions to date, it represents an interesting challenge to NLP systems due to its imbalanced and high number of classes, and some observed inconsistencies in the annotations \cite{fraser2019extracting}. There are  3513 documents (162,908 entities) in the st21pv train set and 879 documents (40,101 entities) in the test set. 

MedMentions was annotated by a team of professional annotators with rich experience in biomedical content curation. The precision of the annotation in MedMention is estimated as 97.3\% \cite{Mohan2019}.  


\paragraph{Model Structures:} We explore a variety of NER deep learning models. For all the models we follow the commonly used deep learning structure consisting of a pretrained embedding model and  supervised prediction layers. For embedding, we explore three different models: a non-contextualized embedding model (Glove), general domain contextualized embedding model (BERT pretrained on general domain text) and a domain-specific contextualized embedding model (BERT pretrained on domain-specific text corpora). For the i2b2 dataset, we consider \textit{Glove+bi-LSTM+CRF} \cite{pennington2014glove}, \textit{BERT+linear} \cite{devlin2018bert} and \textit{ClinicalBERT+linear} \cite{alsentzer-etal-2019-publicly} models. For the st21pv MedMentions dataset, we consider \textit{Glove+bi-LSTM+CRF}, \textit{BERT+linear} and \textit{BioBERT+linear} models \cite{10.1093/bioinformatics/btz682}. Clinical BERT is pretrained on clinical notes (similar to i2b2) and BioBERT is pretrained on biomedical articles from PubMed (similar to st21pv).

\section{Analysis of Error Types Across Models and Datasets}
\label{sec:error-analysis}
\begin{figure*}[h]
\centering
\begin{subfigure}{.5\textwidth}
  \centering
  \includegraphics[width = 8cm]{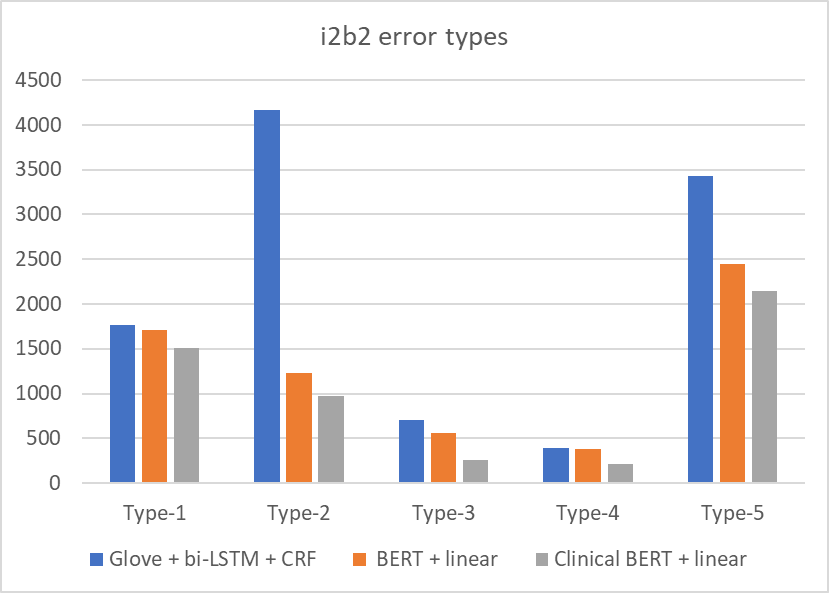}
  \label{fig:errors_i2b2}
\end{subfigure}%
\begin{subfigure}{.5\textwidth}
  \centering
  \includegraphics[width = 8cm]{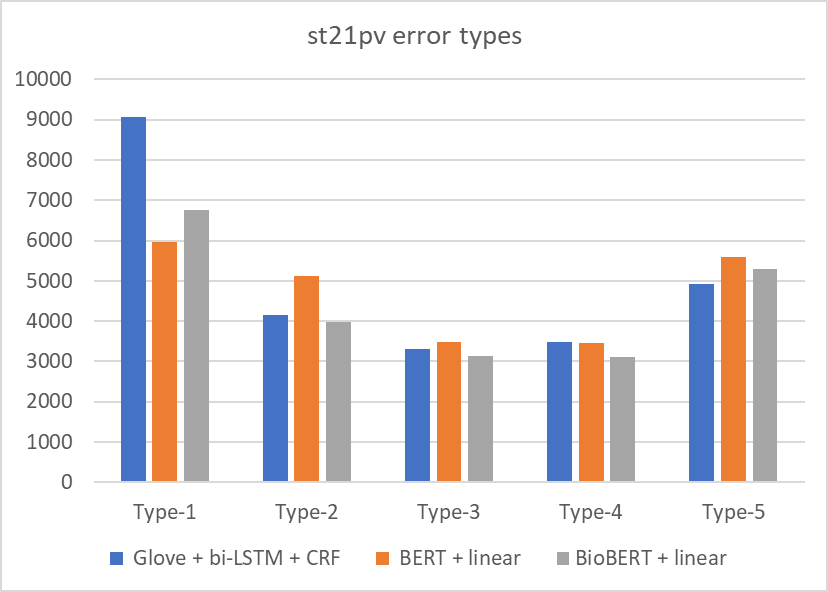}
  \label{fig:errors_stpv}
\end{subfigure}
\caption{Types of errors made on the i2b2 and MedMentions-st21pv datasets}
\label{fig:errors}
\end{figure*}

Further investigation of Type-5 errors is only worthwhile if a significant proportion of the errors belong to this group. We looked at the distribution of error types across datasets and NER models, described in Section \ref{sec:models}, and visualized the results in Figure \ref{fig:errors}. By calculating the distribution of error types, we observed that for all assessed models at least 20\% of the errors are recognized as Type-5 mismatches. 

\begin{figure*}[h]
\centering
\includegraphics[width = 10cm]{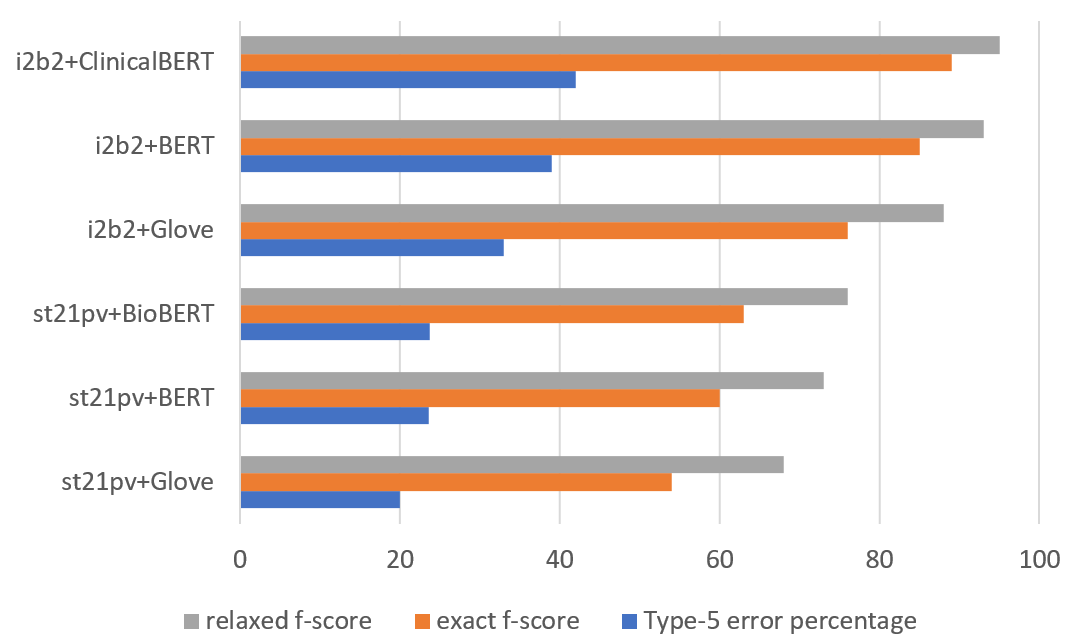}
\caption{The change of relative proportion of Type-5 errors across dataset and models as the f-scores change }
\label{fig:Type-5-percent}
\end{figure*}

Moreover, for both datasets, we observed that better NER models generate more Type-5 errors. Models based on general BERT outperform glove-based models in terms of both exact and relaxed f-score and they also generate relatively more Type-5 errors. Same pattern is observed when comparing domain-specific BERT models with general BERT models. This observation may be explained with the fact that contextualized embeddings combine the meaning of words through attention mechanism and the span information might be more vague in the resulting representation. Figure \ref{fig:Type-5-percent} shows exact F-score, relaxed F-score and the proportion of Type-5 mismatches to the total number of errors, for all the models and datasets. This analysis implies that proper handling of Type-5 errors becomes more important for comparison of modern strong NER systems.

\section{Expert Judgement on Type-5 Errors}
\label{sec:annotation}

We considered an information extraction task and asked a medical doctor to assess the Type-5 errors made by the BioBERT NER model on the st21pv dataset and either confirm or reject the extracted entity with granular scores. Our goal is to: 1) investigate the proportion of Type-5 extracted entities that are acceptable, 2) set a benchmark of human experience from Type-5 errors. 

\paragraph{Human Judgement Scheme:}
The following scoring scheme is used by the expert for scoring the acceptability of Type-5 mismatches for the BioBERT-based model trained with the st21vp dataset. The Type-5 mismatches are identified and the expert is given the original sentence in the test set, the annotated (gold-standard) entity, and the entity predicted by the NER model for all 5296 Type-5 mismatches.  \\ \\
\textbf{SCORE = 1:} The predicted entity is wrong and gets rejected. For example, while \textit{gene transfer} is annotated as a \textit{research\textunderscore activity} in the test set, the NER extracted \textit{gene} as \textit{research\textunderscore activity}. \\ 
\textbf{SCORE = 2:} The predicted entity is correct but an important piece of information is missing when seen in the full sentence. The prediction is partially accepted by the expert. For example, \textit{injury of lung} is labeled as \textit{injury\textunderscore or\textunderscore poisoning} in the test set, but the NER extracts only the word \textit{injury} as \textit{injury\textunderscore or\textunderscore poisoning}.\\ 
\textbf{SCORE = 3:} The predicted entity is correct but could be more complete. The prediction is accepted by the expert. The entity \textit{normal HaCaT lines} is annotated as \textit{anatomical\textunderscore structure} in the test set but the NER extracts only \textit{HaCaT lines} with the same label.  \\ 
\textbf{SCORE = 4:} The predicted entity is equally correct and is accepted by the expert. As an example the annotated entity in test set is \textit{196b-5p,} as an \textit{anatomical\textunderscore structure} but the NER extracts \textit{-196b-5p,} as an entity with the same tag.\\ 
\textbf{SCORE = 5:} The predicted entity is more complete than the annotated entity and is accepted by the expert. The annotated entity in the test set is \textit{drugs} with the tag \textit{chemical} and the NER extracts \textit{Alzheimer's drugs} with the same tag.  \\ 

\begin{figure}[h!]
\centering
\includegraphics[trim={0.5cm 0.1cm 0.5cm 0.1cm},clip,width=1\linewidth]{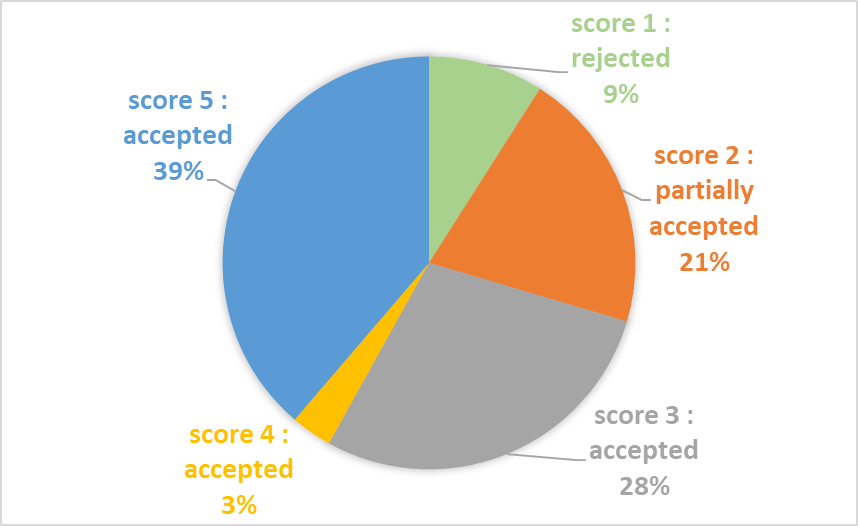}
\caption{Results of expert judgement for Type-5 mismatches of the BioBERT-based NER model trained with MedMentions-st21pv dataset.}
\label{fig:annotation-results}
\end{figure}

\paragraph{Results of Human Judgement Analysis:} The results of the expert judgement are summarized in Figure \ref{fig:annotation-results}. 

\begin{itemize}
    \item Almost 40\% of the Type-5 errors are scored as 5. This means that in 40\% of the cases the prediction of the NER is more complete than the entity labeled in its test set. 
    \item 70\% of the extracted entities scored 3 or above and are fully accepted by the expert.
    \item 21\% of the Type-5 mismatches are scored as 2. These are accepted as a correct entity extraction when seen out of the context, but in the context of a given sentence they lack an important piece of information. Depending on the downstream tasks, they might be an acceptable prediction or not. 
    \item Only 9\% of the extracted entities are totally rejected by the expert. 
\end{itemize}

\section{Entity Classifier for Automatic Refining of Type-5 Mismatches}
\label{sec:classifier}

\begin{figure*}
\centering
\begin{subfigure}{0.8\textwidth}
  \centering
  \includegraphics[width = 8cm]{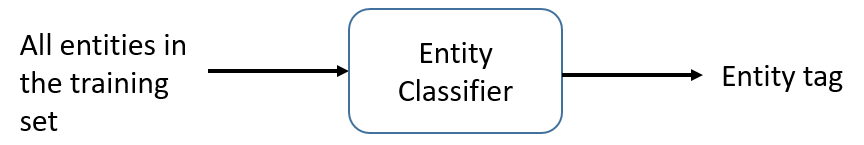}
  \caption{Training}
  \label{fig:BD-training}
\end{subfigure}
\begin{subfigure}{0.8\textwidth}
  \centering
  \includegraphics[width = 11cm]{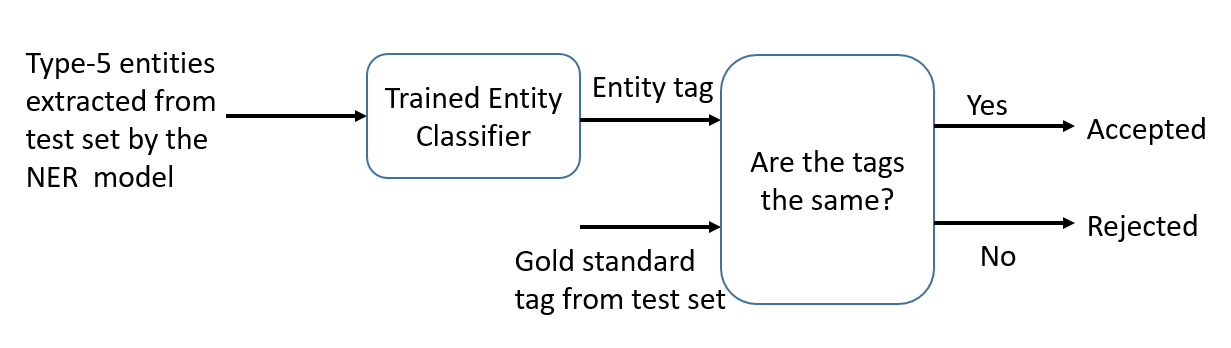}
  \caption{Using for evaluation of the NER model}
  \label{fig:BD-inference}
\end{subfigure}
\caption{Workflow of the proposed entity classifier }
\label{fig:block-diagram}
\end{figure*}

We propose that an entity classifier can be trained to predict the tag of entities extracted by the NER model and the predicted tag can be used to distinguish between acceptable and unacceptable Type-5 errors. Figure \ref{fig:block-diagram} shows the workflow of the proposed method. Using the training dataset of the NER model, we train an entity classifier with gold standard entities as inputs and their assigned tags as outputs. For this classifier, the span is given and the tag is the only information that has to be learned. Although the full context of the sentence helps the NER model to learn a better representation of the entity, many entities can be classified without seeing the full sentence and this is what the entity classifier learns. 


For Type-5 entities, the human annotators and the NER already agree on the tag and it is only the span that is in disagreement. So, the intuition here is that the entity classifier can confirm or reject the tag predicted by NER, given the identified span. This classifier is meant to play a third party role that has seen the variety of span annotations in the training dataset and performs the task that the human expert did in Section \ref{sec:annotation}. This classifier is trained once for each dataset and is not dependent on the type of the NER model. 
 
\subsection{Building the Training Data for the Entity Classifier  }
\label{subsec:training_data}
In order to build a training dataset for the entity classifier, we extracted pairs of \textit{(entity, tag)} from the IOB annotated dataset. The entity classifier should also be able to identify cases where the extracted entity does not belong to any of the pre-defined tags. For this reason we add the label \textit{other} to the list of tags of the classifier. To find examples of the \textit{other} class, we used the spaCy library \cite{spacy2} to extract all the noun chunks that are out of the boundaries of tagged entities and randomly chose a number of them. We limited the size of the \textit{other} class to the average size of classes related to the existing tags.




\subsection{Classifier Structure}
\label{subsec:structure}
For the classifier structure, we chose to use a DistilBERT model \cite{sanh2019distilbert} with a linear prediction layer. DistilBERT is a distilled version of BERT that is an optimum choice when fast inference is required. Since this classifier is going to be used for evaluation and error analysis and is not the main focus of building an NER model, the lightweight and fast inference is an important practical criterion. We train the classifier only one epoch for both datasets. When trained on the train set and tested on the test set, we achieved 89\% F-score for i2b2 and 77\% F-score for st21pv dataset. 

\begin{figure}[h!]
\centering
\includegraphics[trim={0.5cm 2cm 0.5cm 0.1cm},clip,width=1\linewidth]{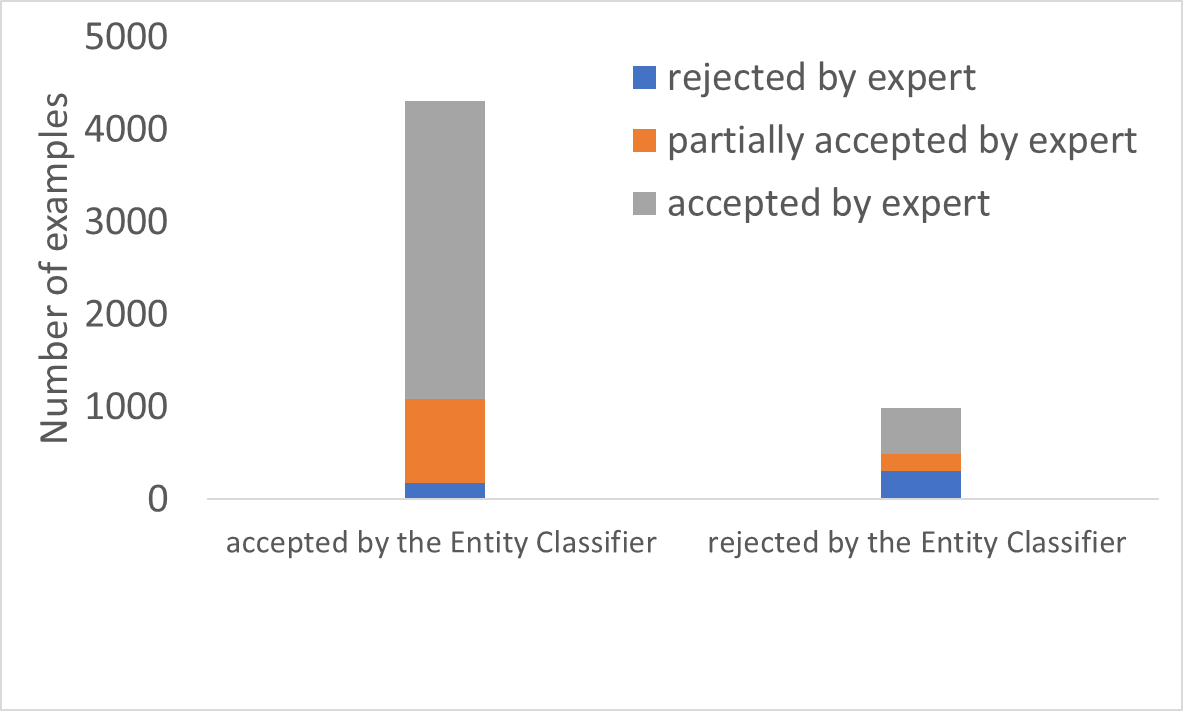}
\caption{Comparison of decisions made by the human expert and the entity classifier for the Type-5 mismatches of BioBERT NER and st21pv dataset. }
\label{fig:compare}
\end{figure}

\begin{table*}
\centering
\begin{tabular}{ccccc}

 {\small \textbf{Annotated in test set}} &  {\small \textbf{Tag in test set}} &  {\small \textbf{Extracted by NER}} &  {\small\textbf{ Tag from Entity classifier}}&{\small \textbf{Decision} } \\
\hline
{\small Central pathology} & {\small biomedical discipline} & {\small Central} & {\small Spatial concept} & {\small Reject}\\
{\small Therapies} & {\small healthcare activity} & {\small Agonist Therapies} & {\small healthcare activity} &{\small Accept} \\

\hline
\end{tabular}
\caption{\label{tab:classifier-decision}
Examples of accepted and rejected Type-5 mismatches using the entity classifier (st21pv dataset).
}
\end{table*}

\subsection{Using the Entity Classifier for Refining Type-5 Mismatches}
\label{subsec:classifier-inference}
By building the entity classifier, our goal is to refine the Type-5 errors and separate the acceptable predictions of the NER from the unacceptable. For instance, in the last example shown in Figure  \ref{fig:examples} there are two Type-5 errors. We feed the two extracted entities \textit{`1 cm cyst in the right lobe'} and \textit{`liver'} to the entity classifier trained for i2b2 dataset. The classifier predicts the tag \textit{`problem'} for the extracted entity \textit{`1 cm cyst in the right lobe'} and \textit{`Other'} for the extracted entity \textit{`liver'}. Using these predictions we decide that the first entity is acceptable, since although the span of the extracted entity does not match the annotation, the classifier still recognizes it as a member of the correct class. We reject the extracted entity \textit{`liver'} as a \textit{`problem'} since the classifier recognizes it as not being a \textit{`problem'}.  Table \ref{tab:classifier-decision} shows examples of rejected and accepted Type-5 mismatches from the st21pv dataset. 

\begin{figure*}
\centering
\includegraphics[width = 10cm]{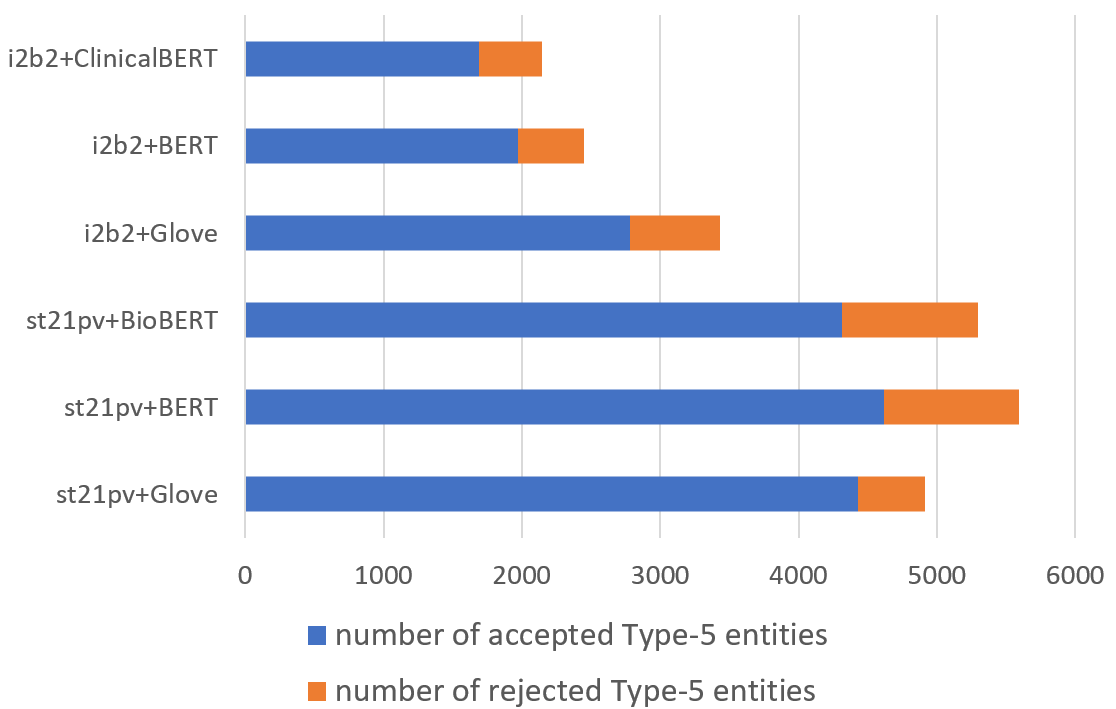}
\caption{Number of accepted/rejected Type-5 mismatches by the entity classifier}
\label{fig:refined-errors}
\end{figure*}

\subsection{Comparing the Classifier and the Expert}
\label{subsec:classifier_evaluation}
Figure \ref{fig:compare} shows the comparison between the expert's judgment and the classifier's judgement about Type-5 mismatches for the BioBERT NER model on st21 pv dataset.

Our analysis shows that 96\% of the entities accepted by the classifier are also accepted or partially accepted by the expert, and 86\% of the entities accepted or partially accepted by the expert are accepted by the classifier as well. The classifier and the expert disagree about 17\% of the entities. In 24\% of the disagreements, the probabilities assigned to the tags generated by the entity classifier are low (less than 0.5) and our manual investigation shows that the classifier's prediction is mostly wrong in these cases. These mistakes mostly occurs in 5 classes namely \textit{anatomical\textunderscore structure}, \textit{biologic\textunderscore function}, \textit{chemical}, \textit{finding} and \textit{health\textunderscore care\textunderscore activity}. 

We also observed that this classifier is not able to distinguish between accepted and partially accepted entities extracted by the NER model, which is one of the limitations of this method. The probabilities assigned to the tags is $0.89\pm0.17$ for accepted entities, $0.88\pm0.17$ for partially accepted entities, and  $0.78\pm0.23$ for rejected entities.

\section{Refining Type-5 Mismatches Across Datasets and Models}

Figure \ref{fig:refined-errors} shows how the entity classifier refines Type-5 errors across models and datasets. Consistently, a significant proportion of Type-5 errors are accepted by the entity classifier. For example, for the Glove-based model trained on i2b2 dataset, the entity classifier accepts 90\% of Type-5 errors which is 26.6\% of the total number of the errors penalized by the exact f-score. The proportion of accepted Type-5 mismatches to the total number of errors is 31.11\% for \textit{i2b2+BERT}, 33.23\% for \textit{i2b2+ClinicalBERT}, 17.95\% for \textit{st21pv+Glove}, 19.55\% for \textit{st21pv+BERT} and 19.36\% for \textit{st21pv+BioBERT}. To sum up, about 20\% to 30\% of the mismatches penalized by exact f-score are accepted by the entity classifier.

\section{Learning-Based F-score}
\label{sec:f-score}


The trained entity classifier can be leveraged for F-score calculation. Here, instead of penalizing all the type-5 mismatches as in exact F-score or rewarding all of them in relaxed F-score, we penalize the type-5 mismatches that are rejected by the classifier and reward the rest of them. In other words, this F-score penalizes errors of Type-1, Type-2, Type-3, Type-4 and the Rejected Type-5 mismatches. Accepted Type-5 mismatches and exact matches are rewarded.  

\subsection{Evaluation of the Learning-Based F-score}
\label{subsec:fscore_eval}

We use the expert judgement collected in Section \ref{sec:annotation} to quantify human experience for the  BioBERT-based NER model on st21pv dataset and then use that as a benchmark to evaluate the proposed learning-based F-score. We consider two scenarios based on the scores described in Section \ref{sec:annotation}, 1) a strict user that only accepts scores equal to or above 3, 2) a forgiving user that accepts scores equal to or above 2. We calculated the F-score for each scenario and investigated the error of exact F-score, relaxed F-score and the proposed F-score to each of these scenarios. Table \ref{tab:f-score-eval} shows that in applications where strict evaluation of the NER is needed, exact F-score is better than both proposed and relaxed f-score and results in the least error with respect to the human experience. However, in cases that partially accepted entities can be considered as useful predictions, the proposed method results in the least disagreement with human experience. A better classifier would be able to model human preferences better, and thus make the learning-based F-score a stronger alternative to exact or relaxed F-scores. Another important finding from Table \ref{tab:f-score-eval} that when choosing between exact and relaxed F-score, exact is the better metric to choose.

\begin{table}
\centering
\begin{tabular}{lcc}
\hline
{\small F-score } & {\small err. wtr strict user} &{\small err. wtr forgiving user}\\
\hline
{\small Exact}  &\textbf{-4.3\%}         & -5.5\%\\ 
{\small Proposed}  & 5.9\%       & \textbf{4.7\%}\\ 
{\small Relaxed}  & 8.2\% & 7.1\% \\ \hline
\end{tabular}

\caption{Comparing F-scores with human experience. }\label{tab:f-score-eval}
\end{table}

\begin{figure*}
\centering
\begin{subfigure}{.5\textwidth}
  \centering
  \includegraphics[width = 8cm]{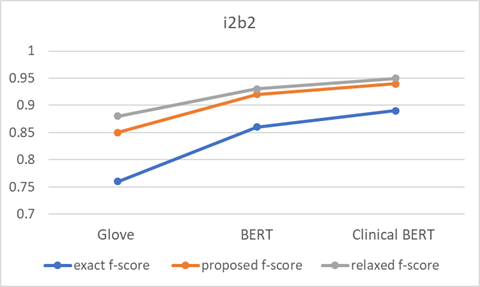}
  \label{fig:i2b2-metrics}
\end{subfigure}%
\begin{subfigure}{.5\textwidth}
  \centering
  \includegraphics[width = 8cm]{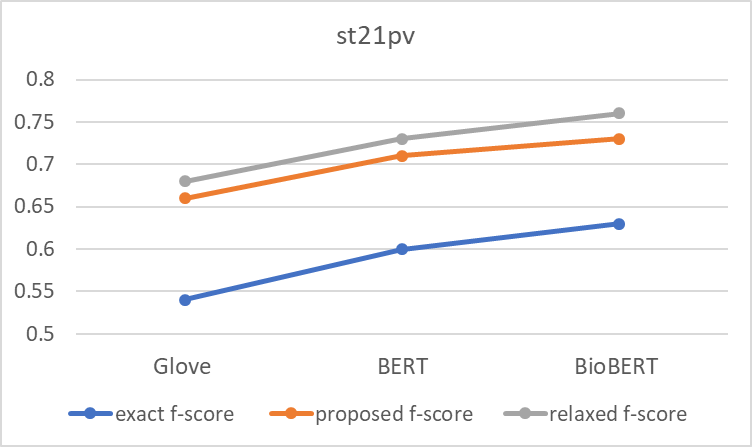}
  \label{fig:stpv-metrics}
\end{subfigure}
\caption{Comparison of f-scores. }
\label{fig:f-scores}
\end{figure*}

Figure \ref{fig:f-scores} shows how the proposed F-score can be compared with exact and relaxed F-score. We only have annotations for the BioBERT+stpv dataset and  for the rest of the models we cannot evaluate the F-score with respect to human experience. As expected, from this figure we observe that for all the models, the proposed F-score is a forgiving one and is much closer to the relaxed F-score than the exact F-score.  

\section{Discussion}
We highlighted the fact that when we evaluate NER systems by comparing extracted and annotated entities across a test set, for a significant part of the errors that are penalized by the exact F-score, the label is recognized correctly and the span has overlap with the annotated entity. We referred to this type of error as Type-5 mismatch and for six NER models (3 model structures and 2 datasets) showed that at least 20\% of the errors belong to this category. The previous literature has raised the issue that in the case of medical NER, many such predictions are valid and useful entity extractions and penalizing them is a flaw of evaluation metrics. However, distinguishing between acceptable and unacceptable predictions when the label is correct and the span overlaps is not trivial.  

We argue that the best evaluation metric is the one that reflects the human experience of the system best. We collected human judgement about a all Type-5 errors made by a NER model based on BioBERT embeddings, trained with st21pv dataset and showed that almost 70\% of such errors are completely acceptable and only 10\% of them are rejected by the user. The rest of the predictions are acceptable entities for the associated tags but lack important information when seen in the context. 

Setting human experience as a benchmark, we suggested that expert judgement can be approximated by a decision made by an entity classifier. The entity classifier can be trained using the training set of an NER. While the NER model looks at the context and identifies the type of the entity and a partially correct span, this classifier looks at the extracted entities out of context and decides whether with the partially correct span, the extracted entity can still belong to the predicted class or not. The entity classifier trained on st21pv dataset accepts more than 80\% of Type-5 errors made by BioBERT-based NER model trained with the same dataset, 96\% of which is also accepted by the expert user. 
The proposed entity classifier is trained for each NER training set once and can be used to evaluate any NER model trained on that dataset, regardless of the structure of the NER model being evaluated. We used a computationally inexpensive model structure and encourage researchers to use this model in order to automatically evaluate Type-5 mismatches. Reporting the distribution of errors across all error types and also accepted and rejected Type-5 errors, will allow us to compare our models in a variety of dimensions and sheds light on how these models behave differently for detecting labels and spans. 

Accepting some Type-5 errors as useful predictions can be translated to F-score calculation by not penalizing the accepted entity extractions. We did this calculation separately for the cases that were accepted by human expert or the classifier, and showed that the F-score resulting from the classifier is closer to the judgement of a forgiving user than both the exact and the relaxed F-score. In cases where a strict evaluation of the system is desired, exact F-score is a better approximation of human experience, due to the fact that the entity classifier is a forgiving one and accepts most of the cases that are partially accepted by the expert. 

We only collected human judgement on the decisions made by NER model for one model and one dataset. Further investigation is needed to confirm or reject our observations and to investigate the limitations and potential capabilities of training an entity classifier alongside a NER model and using that for error analysis. Also, further research is needed to find a way of distinguishing between partially accepted and accepted entity extractions, which is a necessary tool for measuring the experience of a strict user. Using extra sources of training data other than the NER training dataset may be a way to improve the judgements of the entity classifier. We used this classifier for error analysis and refining of Type-5 errors. In future work, we can look at the possibility of using this classifier as a refining tool for all types of mismatches or a post-processing tool without the need for annotation to identify the types of mismatches.

\section{Conclusion}
Medical NER systems that are based on most recent deep learning structures generate a high amount of outputs that match with the hand-labelled entities in terms of tag but only overlap in the span. While the exact f-score penalizes all of these predictions and relaxed f-score credits all of them, a human user accepts a significant proportion of them as valid entities and rejects the rest. \\
A reformatted version of the NER training dataset can be used to train an entity classifier for evaluation of extracted entities with right label and overlapping span. We showed that there is a high degree of agreement between human expert and this entity classifier in accepting or rejecting span mismatches. This classifier is used to calculate a learning-based evaluation metric that outperforms relaxed F-score in approximating the experience of a forgiving user.

\bibliography{acl2020}
\bibliographystyle{acl_natbib}

\end{document}